\documentclass[letterpaper, 10 pt, conference]{ieeeconf}  

\IEEEoverridecommandlockouts                              
\usepackage[utf8]{inputenc}
\usepackage[T1]{fontenc}
\usepackage{graphicx}
\usepackage{booktabs}
\usepackage{float}
\usepackage{mathtools}
\usepackage{amsmath} 
\usepackage{breqn}
\usepackage{caption} 
\title{\LARGE \bf
DCT-HistoTransformer: Efficient Lightweight Vision Transformer with DCT Integration for histopathological image analysis
}

\author{Mahdi Cherakhloo$^{1}$, Mahtab Ranjbar$^{2}$$^*$, Mehdi Mohebbi$^{2}$$^*$ and Bijan Vosoughi. Vahdat$^{1}$
\thanks{$^{1}$Department of Medical Engineering, Electrical Engineering Department, Sharif University of Technology}
\thanks{$^{2}$Department of Mathematical and Computer Sciences, Kharazmi University}%
}

\begin{document}

\maketitle
\def\thefootnote{*}\footnotetext{These authors contributed equally to this work}\def\thefootnote{\arabic{footnote}}
\thispagestyle{empty}
\pagestyle{empty}

\begin{abstract}
In recent years, the integration of advanced imaging techniques and deep learning methods has significantly advanced computer-aided diagnosis (CAD) systems for breast cancer detection and classification. Transformers, which have shown great promise in computer vision, are now being applied to medical image analysis. However, their application to histopathological images presents challenges due to the need for extensive manual annotations of whole-slide images (WSIs), as these models require large amounts of data to work effectively, which is costly and time-consuming. Furthermore, the quadratic computational cost of Vision Transformers (ViTs) is particularly prohibitive for large, high-resolution histopathological images, especially on edge devices with limited computational resources. In this study, we introduce a novel lightweight breast cancer classification approach using transformers that operates effectively without large datasets. By incorporating parallel processing pathways for Discrete Cosine Transform (DCT) Attention and MobileConv, we convert image data from the spatial domain to the frequency domain to utilize the benefits such as filtering out high frequencies in the image, which reduces computational cost. This demonstrates the potential of our approach to improve breast cancer classification in histopathological images, offering a more efficient solution with reduced reliance on extensive annotated datasets. Our proposed model achieves an accuracy of 96.00\% ± 0.48\% for binary classification and 87.85\% ± 0.93\% for multiclass classification, which is comparable to state-of-the-art models while significantly reducing computational costs. This demonstrates the potential of our approach to improve breast cancer classification in histopathological images, offering a more efficient solution with reduced reliance on extensive annotated datasets.\\
\end{abstract}


\section{INTRODUCTION}

Automated detection and classification of breast cancer from histopathological images plays a pivotal role in the clinical diagnosis and prognosis within the framework of the Precision Medicine Initiative \cite{a1}\cite{a2}. The World Cancer Report issued by the World Health Organization delineates breast cancer as the predominant neoplasm among women globally, with unparalleled morbidity and mortality rates. Statistically, breast cancer constitutes 25.2\% of cancer incidences among female patients, occupying the foremost rank, whereas its morbidity rate stands at 14.7\%, securing the second rank subsequent to lung cancer in recent epidemiological studies on cancer mortality. Annually, approximately half a million fatalities are attributed to breast cancer, alongside the emergence of nearly 1.7 million new cases, a figure anticipated to escalate markedly. Moreover, histopathologic imagery is acclaimed as the quintessential standard over alternative medical imaging techniques such as mammography, magnetic resonance imaging (MRI), and computed tomography (CT) for the delineation of breast cancer. Notably, the determination of an optimal therapeutic regimen for breast cancer hinges significantly on advanced multi-classification processes. This is attributed to the fact that clinicians, armed with knowledge of the specific subclasses of breast cancer, can promptly intervene in the metastasis of tumor cells, thereby devising tailored therapeutic strategies based on the unique clinical manifestations and prognostic outcomes of various breast cancer subtypes.\\
Indeed, conducting manual classification on breast cancer histopathological images poses significant challenges. The challenges stem from three primary factors: (1) the difficulty in passing down or enhancing the specialized knowledge and extensive experience of pathologists, which results in a scarcity of proficient pathologists in entry-level hospitals and clinics, (2) the laborious nature of the task, which is both costly and time-intensive, and (3) the risk of misdiagnosis due to pathologist fatigue. Therefore, there is a critical and pressing need for implementing computer-aided multi-classification for breast cancer, which could significantly alleviate the workloads of pathologists and minimize the risk of diagnostic errors \cite{a3}.\\
Over the past decade, deep learning methodologies have elicited a paradigm shift across multiple sectors, notably within healthcare, by facilitating tasks such as precise disease diagnosis, prognostication, and the advancement of robotic-assisted surgical procedures. Research leveraging deep convolutional neural networks (CNNs) has been conducted to detect breast cancer (BC) utilizing the medical imaging techniques aforementioned. Nevertheless, CNNs are predisposed to innate inductive biases, exhibiting sensitivity to the translation, rotation, and spatial positioning of the target object within images. Consequently, image augmentation techniques are commonly employed during the training of CNN models to introduce variability, although these techniques may not always yield the anticipated diversity within the training dataset. In response to these limitations, there is an emergent focus on the development of deep learning models predicated on self-attention mechanisms. These models demonstrate enhanced resilience to variations in the orientation and location of the object of interest within images, thereby representing a significant advancement in the field \cite{a4}.\\
In recent years, ViTs \cite{a5}, have emerged as a powerful alternative to traditional Convolutional Neural Networks (CNNs) in computer vision tasks such as image classification \cite{a13}\cite{a14}, object detection \cite{a15}\cite{a16}, and image segmentation \cite{a17}\cite{a18}. The core component of Transformer is the attention  mechanism, which computes dependencies between all pairs of tokens in a sequence. However, for a sequence of length $L$, the expressiveness of pairwise attention comes at a quadratic cost $O(L^{2})$ in both  time and memory consumption. As a result, the computational demands of ViTs become a bottleneck,  hindering their efficient deployment. This problem holds great importance for deep learning  applications tackling high-resolution images and those functioning on edge devices with limited  computational resources.\\
In this paper, we pursued two objectives to address this challenge:\\

\subsection{Reducing Input Data Size for Self-Attention and Lower Computational Cost}

The first objective of this research is to leverage the concept of Frequency Transformers to transform  spatial data into the frequency domain. This transformation allows us to remove high-frequency components, effectively reducing the size of the input. By decreasing the input size, we can alleviate the computational demands associated with self-attention weight calculations. The novelty of this objective  lies in the application of frequency-based transformations for reducing input data size which has the potential to significantly reduce computational costs while preserving essential information for accurate predictions. \\

\subsection{Capturing Local Dependencies to Maintain Performance while Reducing Input Size and Computational Cost}

The second objective is to tackle the potential loss of local interactions that may occur when reducing  the input size. To overcome this challenge, we propose incorporating MobileBlock convolutions into the architecture. We aim to preserve the ability of the model to capture fine-grained details and local patterns in the data. This objective ensures that the proposed model maintains its capability to handle local interactions while benefiting from the computational efficiency achieved through the reduction of input size.\\\\
Our proposed DCT-Conv block places two parallel branches at the forefront - DCT-Attention and MobileConv, each providing independent preprocessing to raw image data. Figure 2 shows the overall  architecture.\\

\section{RELATED WORKS}
Vision Transformers (ViTs) were first introduced by Dosovitskiy et al.\cite{a5}, marking the beginning of a new era in image recognition tasks. However, these initial ViTs were associated with high computational costs and quadratic complexity for both time and memory due to their global self-attention mechanism. As a result, subsequent research efforts shifted towards proposing lightweight ViTs that aimed to strike a balance between performance, computational cost, and speed.\\
One approach has been to tune the transformer part of the model, enhancing self-attention mechanisms and redesigning the Transformer blocks while maintaining lower computational costs. Swin-Transformer \cite{a6} proposed an efficient self-attention mechanism,  employing window-based and shift mechanisms to capture long-range dependencies efficiently. While effective, its high parameter and FLOPS count  hindered its suitability for resource-constrained edge devices and mobile applications. In response, PSLT \cite{a7} proposed ladder self-attention blocks with a progressive shift mechanism which divides the input  feature map into multiple branches, where each branch shifts the obtained features in different directions and divides them into multiple windows for window-based self-attension layer. This enables  the aggregation of output features from diverse windows, allowing for efficient modeling of long-range  interactions. This approach significantly reduces the number of parameters and FLOPS. However, further exploration of alternative computation methods for faster operations is needed, as its slower inference  rate suggests. In order to address the challenges associated with slow inference speeds, EfficientViT \cite{a8}, a high-speed vision transformer family adopts a sandwich layout that improves memory usage, placing  a single memory bound Multi-Head Self-Attention layer between efficient Feed-Forward layers. Furthermore, computation costs are minimized through the integration of a novel cascaded group  attention module that feeds each attention head with different splits of the complete feature. The EfficientViT excels in enhancing computational efficiency and speed, offering a practical solution for real time edge applications. 
Another research approach that aims to address the challenges of lightweight ViTs involves exploring  hybrid models that combine the strengths of convolutions capturing local dependencies and transformers in handling global and long-range dependencies.\cite{a9}\cite{a10}. MOAT \cite{a11} introduces a novel  building block that includes an MLP after the self-attention layer with a mobile convolution, and then reorders and places it before the self-attention layer. By doing so, this approach enhances representation capacity and delegates downsampling to strided depth-wise convolution. However, further optimization is needed considering its relatively high parameter and FLOPS count compared to other state-of-the-art (SOTA) lightweight hybrid models. \\ 
Studies in the field of lightweight models like the ones mentioned earlier have primarily focused on designing delicate network architectures, often neglecting the optimization of training strategies. However, it has been observed that proper pre-training can lead to comparable performance of even vanilla lightweight ViTs compared to more complex designs. In this context, MAE \cite{a12} evaluated self-supervised pre-training methods on ViT-Tiny and proposed MAE-Tiny, which is a vanilla ViT-Tiny model pre-trained using Masked Image Modeling (MAE). MAE-Tiny exhibited limitations, as it struggled to effectively leverage large pre-training datasets and performed poorly on downstream tasks with limited data. To overcome these limitations, DMAE-Tiny was introduced which involved distilling knowledge from a larger MAE model into the smaller MAE-Tiny model during the pre-training phase. It significantly improved results on data-limited classification and detection tasks where MAE-Tiny had previously shown deficiencies. However, these methods do not fully exploit the advantages of large-scale pre-training data. \\
In histopathology applications, several studies have utilized ViTs to improve classification accuracy. Wang et al. \cite{a19} proposed a semi-supervised learning procedure based on ViT, incorporating adaptive token sampling (ATS) and consistency training (CT) strategies. This approach combines supervised and unsupervised learning with image augmentation, achieving an average test accuracy of 0.98 on the BreakHis dataset. Tummala et al. \cite{a20} used a variant called Swin Transformer (SwinT), which uses non-overlapping shifted windows to improve efficiency. Their model is an ensemble of pre-trained SwinT models which includes the tiny, small, base, and large variants. Alotaibi et al. \cite{a21} introduced an ensemble model merging ViT and Data-Efficient Image Transformer (DeiT) Its novelty lies in the addition of DeiT, which uses an extra input called the distillation token. The learning process of the distillation token involves backpropagation, where it interacts with classes and patch tokens through the self-attention layers.\\\\
\raggedbottom 
\section{METHOD}
\subsection{Dataset}
The BreaKHis dataset comprises 7,909 microscopic RGB images derived from the surgical biopsy of breast tumors across 82 patients, captured at magnifications of 50×, 100×, 200×, and 400×. Figure 1 displays representative images from the dataset across these varying levels of magnification. The dataset encompasses both benign and malignant tumor subtypes. Specifically, the benign subtypes consist of fibroadenoma, tubular adenoma, phyllodes tumor, and adenosis, while the malignant categories include ductal carcinoma, papillary carcinoma, lobular carcinoma, and mucinous carcinoma. Table 1 provides detailed information regarding the dataset, organized by tumor type and magnification levels.\\\\

\subsection{Discrete Cosine Transform}
The Discrete Cosine Transform (DCT) is a widely used transformation technique in image processing, primarily known for its ability to compactly represent the energy of a signal in the frequency domain. This characteristic makes the DCT particularly useful for tasks such as image compression and noise reduction. While the DCT is originally defined for one-dimensional (1D) signals, it can be extended to two-dimensional (2D) images to analyze and manipulate spatial information in a more effective manner.

\begin{figure}[t]
    \centering
    \caption{Representative images from the BreaKHis dataset are displayed at magnification levels of 40×, 100×, 200×, and 400×. The abbreviations represent the following conditions: AD corresponds to adenosis, FA to fibroadenoma, PT to phyllodes tumor, TA to tubular adenoma, DC to ductal carcinoma, LC to lobular carcinoma, MC to mucinous carcinoma, and PC to papillary carcinoma.}
    \includegraphics[width=0.5\textwidth]{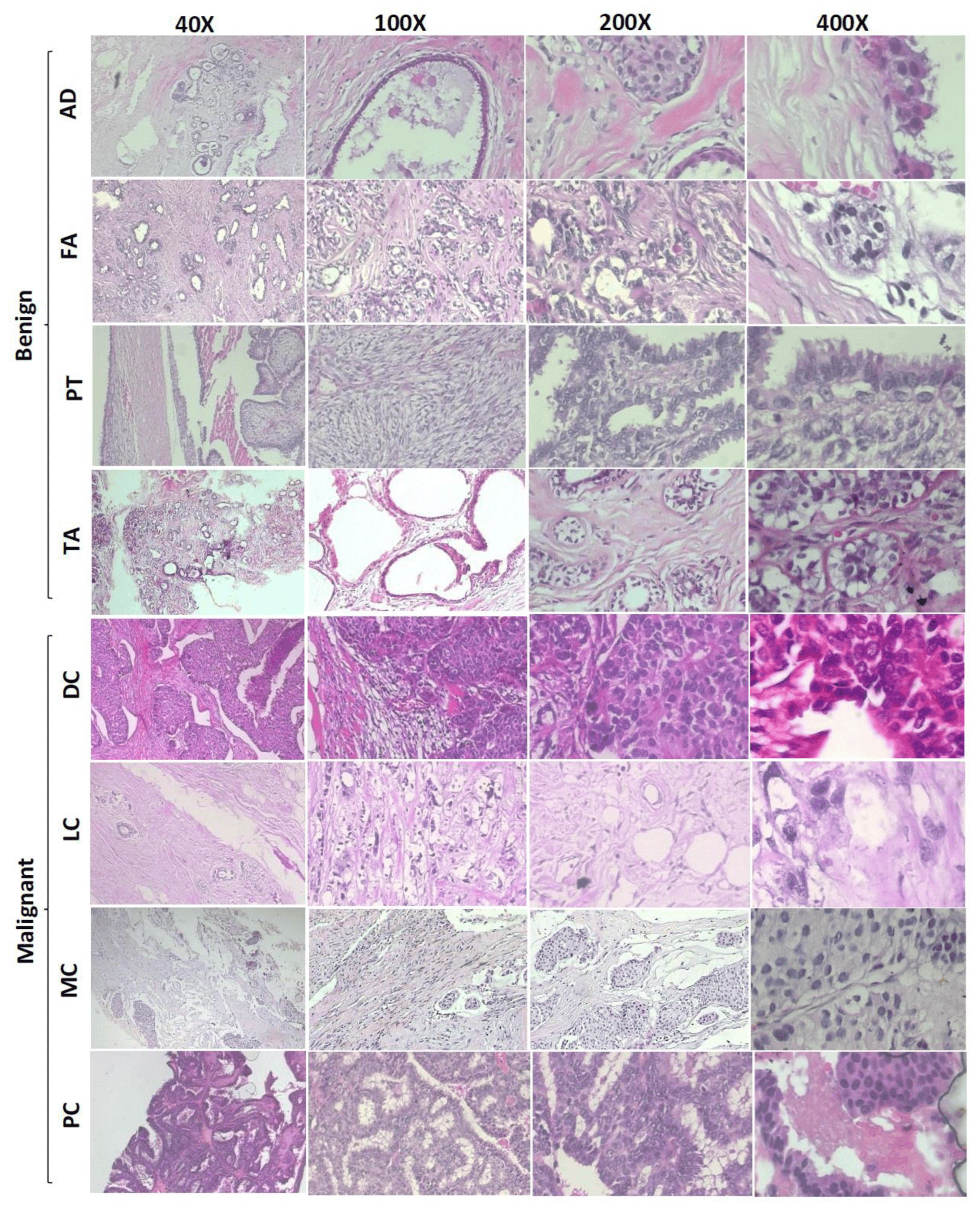}
    
    \label{fig:data}
\end{figure}

The DCT of a one-dimensional signal \( x[n] \) of length \( N \) is defined by the following formula:
\begin{equation}
X[k] = \sum_{n=0}^{N-1} x[n] \cdot \cos \left( \frac{\pi}{N} \left( n + \frac{1}{2} \right)
k \right), \quad k = 0,  \ldots, N-1    
\end{equation}

In this formula, \( X[k] \) represents the DCT coefficient at frequency index \( k \), \( x[n] \) is the input signal, and \( N \) is the length of the input signal.\\

In the two-dimensional (2D) case, the DCT is applied to images as a matrix of pixel values. It extends the 1D DCT to analyze both spatial dimensions of an image. The 2D DCT is defined as:

\begin{dmath}
 X[u, v] = \alpha(u) \alpha(v) \sum_{x=0}^{M-1} \sum_{y=0}^{N-1} x[x, y] \cdot \cos \left[ \frac{\pi}{M} \left(x + \frac{1}{2}\right) u \right] \cdot \cos \left[ \frac{\pi}{N} \left(y + \frac{1}{2}\right) v \right]
\end{dmath}

where:
\begin{itemize}
    \item \(x[x, y]\) denotes the pixel value at position \((x, y)\) in the image.\\
    \item \(X[u, v]\) is the DCT coefficient at frequency \((u, v)\).
    \item \(\alpha(u)\) and \(\alpha(v)\) are normalization factors:
    \begin{align}
    \alpha(u) &= \begin{cases} 
    \sqrt{\frac{1}{M}} & \text{for } u=0, \\
    \sqrt{\frac{2}{M}} & \text{for } u \neq 0 
    \end{cases} \\
    \alpha(v) &= \begin{cases} 
    \sqrt{\frac{1}{N}} & \text{for } v=0, \\
    \sqrt{\frac{2}{N}} & \text{for } v \neq 0 
    \end{cases}
    \end{align}
\end{itemize}

\begin{table}[h]
\centering
\caption{The quantity of images at magnification levels of 40×, 100×, 200×, and 400× from the BreaKHis dataset is categorized according to cancer subtype. The acronyms denote: FA for fibroadenoma, TA for tubular adenoma, PT for phyllodes tumor, AD for adenosis, DC for ductal carcinoma, LC for lobular carcinoma, MC for mucinous carcinoma, and PC for papillary carcinoma. 'N' represents the total number of images within each categorization.}
\begin{tabular}{lccccc}
\toprule
\textbf{Tumor Type} & \textbf{40x} & \textbf{100x} & \textbf{200x} & \textbf{400x} & \textbf{All} \\
\midrule
FA & 253 & 260 & 264 & 237 & 1014 \\
TA & 149 & 150 & 140 & 130 & 569 \\
PT & 109 & 121 & 108 & 115 & 453 \\
AD & 114 & 113 & 111 & 106 & 444 \\
DC & 864 & 903 & 896 & 78  & 1041 \\
PC & 145 & 142 & 135 & 138 & 560 \\
LC & 156 & 170 & 163 & 137 & 626 \\
MC & 205 & 222 & 196 & 169 & 792 \\
\midrule
\textbf{Total (N)} & 1995 & 2081 & 2013 & 1820 & 7909 \\
\bottomrule
\end{tabular}

\label{tab:tumor_type_distribution}
\end{table}

\begin{figure}[h]
    \centering
    \caption{Illustration of DCT-Conv block architecture (In the context of our model H, W, and W refer to the height, width, and the number of channels of the input image, respectively. For our dataset, H = 224, W = 224, and C = 3 representing a 224 x 224 RGB image). (a) MobileConv branch. (b) DCT-Attention branch. (c) Architecture of  Cascaded Group Attention layer.}
    \includegraphics[width=0.5\textwidth]{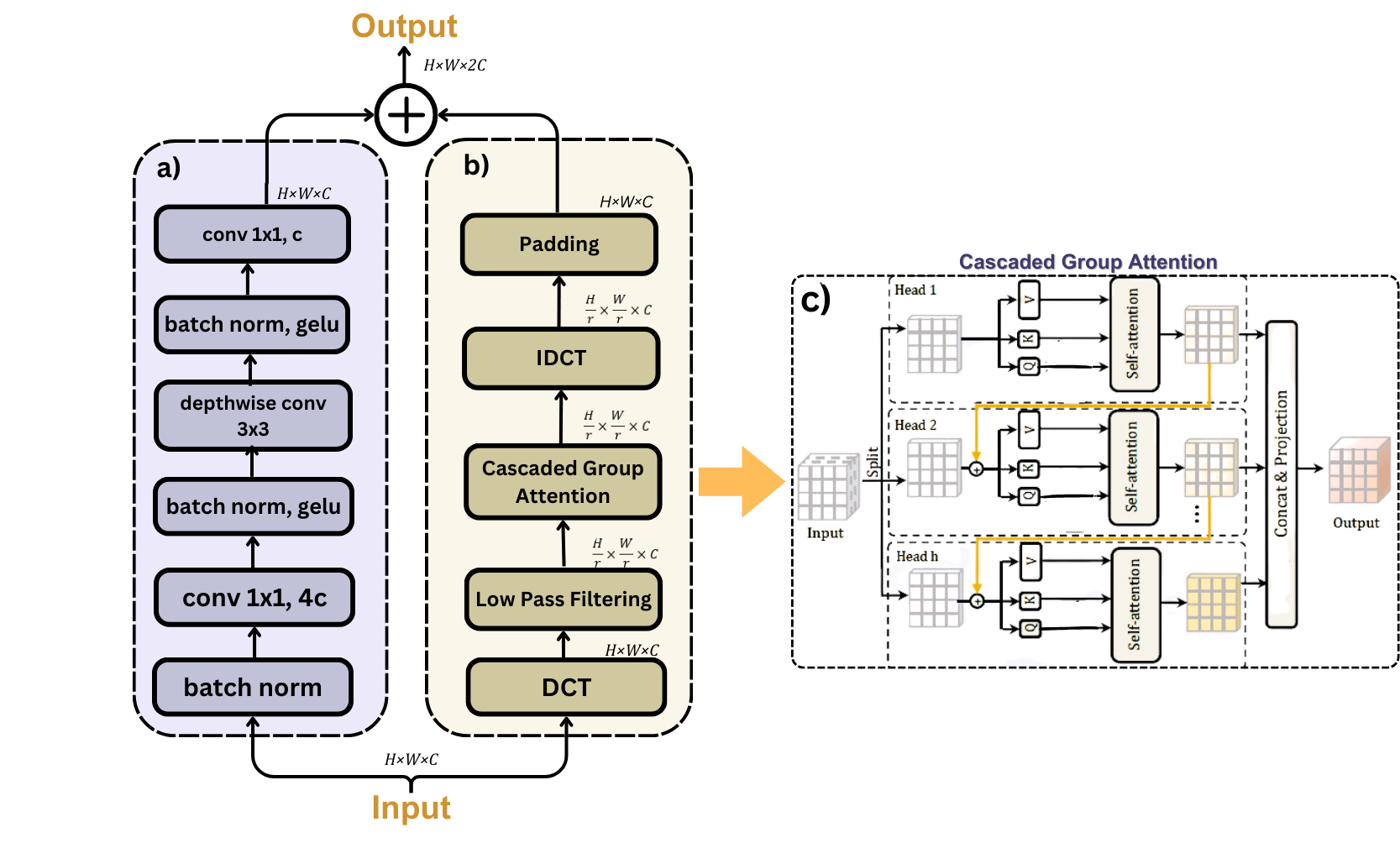}
    
    \label{fig:dct-vit}
\end{figure}

Our proposed DCT-Conv block places two parallel branches at the forefront - DCT-Attention and  MobileConv, each providing independent preprocessing to raw image data. Figure \ref{fig:dct-vit} shows the overall  architecture. \\
\subsection{Methodology-1.} 
DCT-Attention branch: The DCT-Attention branch (Figrue 2 (b)) consists of three key steps: DCT and Low Pass Filtering, Self-Attention with DCT Output, and IDCT. 
First, we apply the Discrete Cosine Transform (DCT) to convert the input image into frequency  components. The DCT separates the image into different frequency bands, where low-frequencies represent important global patterns and high frequencies capture fine details. Then we normalize the DCT coefficients to handle the DC component effectively. This ensures that the DC component, which represents the overall energy of the image, is retained while preventing it from dominating the feature space. By normalizing, we balance the influence of the DC component with other frequency components, preserving essential global information without overemphasis. After the DCT, we perform  a lowpass filtering operation, which scales down the input dimensions by a factor of ”r” (resulting in an  input size of to retain only the low-frequency components and discard the high-frequency components. This filtering helps reduce the input size for attention layer by focusing on preserving only  low-frequency components. Next, Inspired by the Cascade Group Attention layer (CGA) \cite{a8}, we modify  the self-attention mechanism to enhance efficiency. Instead of using the same full feature for all  attention heads, we split the feature into channel-wise partitions. Each attention head receives a  different split, promoting the learning of distinct patterns and reducing computational redundancy. The  attention maps of each head are computed in a cascaded manner, progressively refining the feature  representations (Figure 2 (c)). After the self-attention, we apply the Inverse Discrete Cosine Transform  (IDCT) to reconstruct the spatial representation from the frequency domain. This step bridges the gap  between the frequency based information obtained and the spatial information required for subsequent  processing. Finally, we introduce a padding operation to adjust the size of the image and match it with  the output size of the MobileConv block. This padding ensures compatibility between the feature maps  from the DCT-Attention branch and the subsequent processing in the MobileConv branch. 

\subsection{Methodology-2.} 
MobileConv Block Branch: To capture local dependencies that may be lost during the reduction of input  size after low pass filtering, we introduce a MobileConv branch (Figure 2 (a)) to extract local information  from the raw input data. This branch complements the global patterns learned through the self-attention mechanism by capturing detailed local features, thereby enhancing the overall representation. 
Finally, the outputs from both branches are combined to leverage both the global contextual information obtained from the DCT-Attention branch and the local features captured by the MobileConv branch. This integration allows us to enhance the overall representation by benefiting from the strengths  of both branches. We organize the blocks into different stages within the architecture which allows for  customizing and adjusting the depth and complexity of the model based on the specific needs of the  task or dataset. The stage layout serves as a framework for structuring the model and optimizing its  performance by effectively allocating computational resources and managing the flow of information  through the different stages.

\begin{figure*}[h]
    \centering
    \caption{Overall Architecture of DCT-HistoTransformer}
    \includegraphics[width=0.8\textwidth]{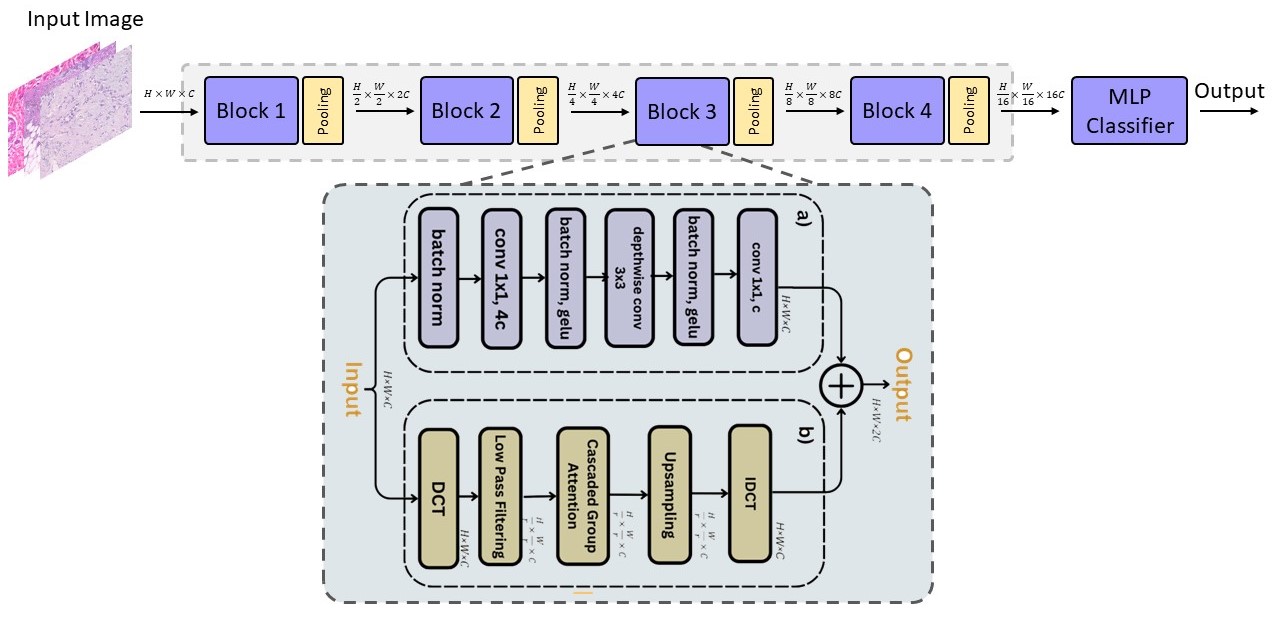}
    
    \label{fig:dct-overall}
\end{figure*}
\section{RESULTS}
\subsection{Evaluation metrics}
Due to the class imbalance present in both datasets, utilizing solely accuracy (Acc) as a measure does not accurately convey the actual effectiveness of a model. Hence, along with Acc, we incorporate additional metrics such as precision (Pre), recall (Rec), and F1 scores for a thorough assessment of performance. The formulation of these metrics is provided in equations 1-4.\\
\begin{equation}
Acc = \frac{TP + TN}{TP + TN + FP + FN}
\end{equation}
\begin{equation}
Pre = \frac{TP}{TP + FP}
\end{equation}
\begin{equation}
Rec = \frac{TP}{TP + FN}
\end{equation}
\begin{equation}
F1-Score = 2\times\frac{Pre \times Rec}{Pre + Rec}
\end{equation}
\\
In this context, TP, TN, FP, and FN denote the counts of true positives, true negatives, false positives, and false negatives, correspondingly. Precision (Pre) measures the proportion of false alarms, with a higher Pre indicating a reduced number of false alarms generated by the model. Recall (Rec), on the other hand, quantifies the number of positive instances that were not detected. Therefore, a higher Rec signifies a decrease in missed positive cases. The F1 score is the harmonic mean of Pre and Rec, offering a metric more appropriate than accuracy (Acc) for evaluating classification tasks when dealing with datasets that are not balanced.\\

\subsection{Training setting}
We evaluated the proposed model on the BreakHis dataset with a batch size of 32. The model’s architecture consists of a total of 7966589 parameters. It includes four blocks, each with MobileConvolution branches and DCT-Attention branches. Figure \ref{fig:dct-overall} shows the model configuration. During training, we applied data augmentation techniques, including RandomHorizontalFlip, RandomRotation, and RandomCrop, to the training images, while the validation images were used without augmentation. The model was trained from scratch using the AdamW optimizer for 50 epochs, with an initial learning rate of 0.0001 and a weight decay of 0.01. The cross-entropy loss function was used to measure performance.

\subsection{Results}

Our proposed model demonstrates a high-performance level, achieving an accuracy of 0.9698 ± 0.0048, a F1-score of 0.9564 ± 0.0049, a precision rate of 0.9565 ± 0.0049, and a specificity rate of 0.9566 ± 0.0049 (Table II, Fig. 4). This level of accuracy is comparable to other state-of-the-art models while significantly reducing computational costs. The model leverages parallel processing pathways in the initial stages, where the DCT-Attention and MobileConv modules operate. This dual-pathway approach transforms the data from the spatial domain to the frequency domain, capturing more global information about the entire image rather than focusing solely on individual pixels.
 In transitioning to the frequency domain, we apply low-pass filtering, which effectively reduces the input size by eliminating high-frequency components typically containing detailed information, noise, and other less important details. This selective retention prioritizes critical low-frequency components, enhancing the recognition and classification accuracy of our model.
 Additionally, the separated local features from the MobileConv branch are integrated back into the final output. This integration further boosts the model’s ability to recognize and classify images accurately across various tasks. The results from Tables IV and V below support these claims.
 As shown in Table III, In multi-class classification tasks, the model achieved an accuracy of 0.8785 ± 0.0093 , a recall of 0.8735 ± 0.0093, a precision of 0.8771 ± 0.0091, and an F1 score of 0.8731 ± 0.0096 .(Fig. 5) These results illustrate the model’s competency in handling more complex classification tasks while maintaining reasonable performance metrics.
 The performance comparison table(Table IV) shows that our model outperforms other existing methods across all magnification levels (40X, 100X, 200X, 400X). Notably, at 200X magnification, our model achieves the highest performance with an accuracy of 97.68\%, illustrating its superior ability to handle different resolutions and magnifications in image classification tasks.
 These comprehensive results highlight the significant advancements our model offers in terms of accuracy and computational efficiency, confirming its effectiveness across a broad range of classification tasks.

\begin{table}[H]
\centering
\caption{Model results for binary classification (malignant vs benign) for 40x magnification images.}
\begin{tabular}{lc}
\toprule
\textbf{Metric} & \textbf{Value} \\
\midrule
Accuracy  & 96.00 ± 0.48\\
Recall    & 95.66 ± 0.49\\
Precision & 95.65 ± 0.49\\
F1 score  & 95.64 ± 0.49\\
\bottomrule
\end{tabular}

\label{tab:model_results}
\end{table}

\begin{table}[H]
\centering
\caption{Model results for multi-class classification for 40x magnification images.}
\begin{tabular}{lc}
\toprule
\textbf{Metric} & \textbf{Value} \\
\midrule
Accuracy  & 87.85 ± 0.93 \\
Recall    & 87.35 ± 0.93\\
Precision & 87.71 ± 0.91\\
F1 score  & 87.31 ± 0.96 \\
\bottomrule
\end{tabular}

\label{tab:model_results_multi}
\end{table}

\begin{table}[ht]
\centering
\caption{Performance comparison of different methods at various magnifications.}
\begin{tabular}{lcccccc}
\toprule
Methods & 40X & 100X & 200X & 400X\\
\midrule
VGG16 \cite{a24} & $90.00 $ & $85.00$ & $85.10 $ & $84.80$\\
RseNet50 \cite{a25} & $92.70$ & $94.40$ & $96.00$ & $95.5$\\
ViT \cite{a5} & $84.10$ & $83.40$ & $91.10$ & $84.10$\\
Swin-Transformer \cite{a6} & $84.30$ & $87.70$ & $87.60$ & $84.60$\\
\textbf{Ours} & $\mathbf{96.00}$ & $\mathbf{95.52}$ & $\mathbf{97.68}$ & $\mathbf{95.23}$ \\
\bottomrule
\end{tabular}

\label{tab:performance_comparison}
\end{table}

\begin{figure}[h]
    \centering
    \caption{ Learning curves of the proposed method on the BreakHis dataset for binary classification (malignant vs benign), showing accuracy and loss over the training epochs.}
    \includegraphics[width=0.5\textwidth]{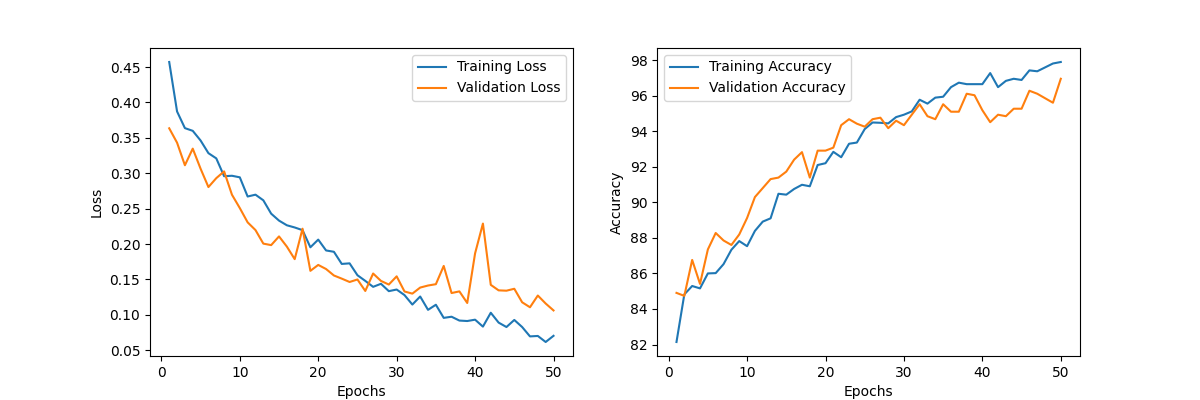}
    \label{fig:curve}
\end{figure}

\begin{figure}[h]
    \centering
    \caption{Learning curves of the proposed method on the BreakHis dataset for multiclass classification, showing accuracy and loss over the training epochs.}
    \includegraphics[width=0.5\textwidth]{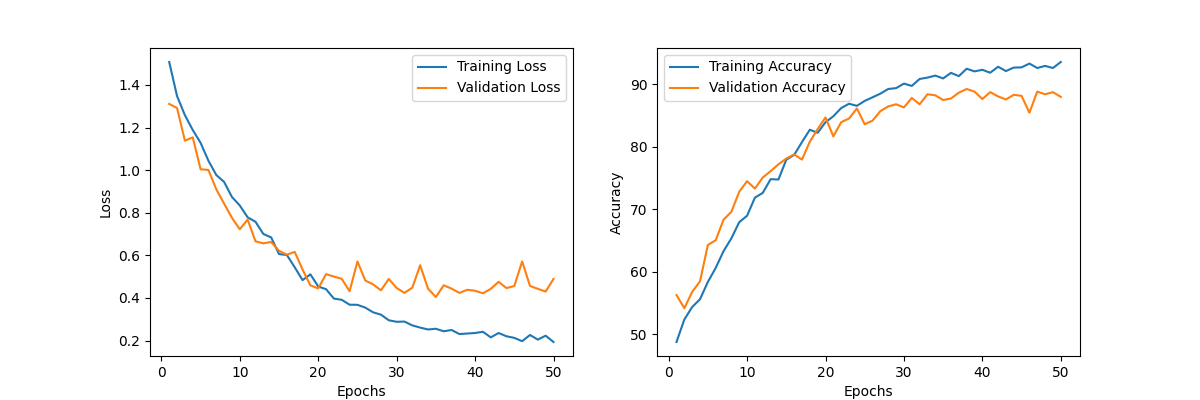}

    \label{fig:curve}
\end{figure}

\section{CONCLUSIONS}
In this study, we have presented a novel approach to breast cancer classification from histopathological images using a lightweight Vision Transformer model. Our approach combines Discrete Cosine Transform (DCT) with self-attention mechanisms to achieve a balance between performance and computational efficiency. By incorporating DCT-Attention and MobileConv branches, our model successfully reduces the computational burden associated with high-resolution images while maintaining a high level of classification accuracy. The effectiveness of our approach lies in its ability to leverage frequency-domain information for efficient data representation and its ability to capture both global and local features through the integration of the DCT-Attention and MobileConv components. Our results highlight the potential of hybrid transformer architectures in the context of medical image analysis, specifically for histopathological image classification.

\section{Future Work}
While our model demonstrates strong performance and efficiency, there remains considerable room for improvement.\\ 
Our current approach uses a simple low-pass filtering mechanism to discard high-frequency components. Future research could investigate more sophisticated methods for selecting DCT coefficients that focus on identifying and preserving those that have the most significant impact on the model’s decision-making process. Techniques such as adaptive thresholding or coefficient pruning based on feature importance could be explored to better balance the trade-off between computational efficiency and model performance.\\
Although the DCT-Attention mechanism has proven effective, there is potential for further refinement of attention mechanisms to improve both the accuracy and efficiency of the model. Exploring advanced attention techniques or hybrid attention mechanisms that integrate DCT with other forms of attention could offer new insights into optimizing the model for specific types of histopathological features.

\addtolength{\textheight}{-12cm}   



\section*{APPENDIX}


\bibliographystyle{ieeetr}
\bibliography{Refrences}

\begin{thebibliography}{10}

\bibitem{a1}
F.~Collins and H.~Varmus, ``A new initiative on precision medicine,'' {\em The New England journal of medicine}, vol.~372, 01 2015.

\bibitem{a2}
S.~Reardon, ``Us precision-medicine proposal sparks questions,'' {\em Nature}, vol.~517, p.~540, 01 2015.

\bibitem{a3}
Y.~Cai, M.~Landis, D.~T. Laidley, A.~Kornecki, A.~Lum, and S.~Li, ``Multi-modal vertebrae recognition using transformed deep convolution network,'' {\em Computerized Medical Imaging and Graphics}, vol.~51, pp.~11--19, 2016.

\bibitem{a4}
T.~I. Alshafeiy, A.~Matich, C.~M. Rochman, and J.~A. Harvey, ``{Advantages and Challenges of Using Breast Biopsy Markers},'' {\em Journal of Breast Imaging}, vol.~4, pp.~78--95, 08 2021.

\bibitem{a5}
A.~Dosovitskiy, L.~Beyer, A.~Kolesnikov, D.~Weissenborn, X.~Zhai, T.~Unterthiner, M.~Dehghani, M.~Minderer, G.~Heigold, S.~Gelly, J.~Uszkoreit, and N.~Houlsby, ``An image is worth 16x16 words: Transformers for image recognition at scale,'' {\em CoRR}, vol.~abs/2010.11929, 2020.

\bibitem{a13}
H.~Touvron, M.~Cord, M.~Douze, F.~Massa, A.~Sablayrolles, and H.~Jégou, ``Training data-efficient image transformers \& distillation through attention,'' 2021.

\bibitem{a14}
K.~Han, A.~Xiao, E.~Wu, J.~Guo, C.~XU, and Y.~Wang, ``Transformer in transformer,'' in {\em Advances in Neural Information Processing Systems} (M.~Ranzato, A.~Beygelzimer, Y.~Dauphin, P.~Liang, and J.~W. Vaughan, eds.), vol.~34, pp.~15908--15919, Curran Associates, Inc., 2021.

\bibitem{a15}
N.~Carion, F.~Massa, G.~Synnaeve, N.~Usunier, A.~Kirillov, and S.~Zagoruyko, ``End-to-end object detection with transformers,'' 2020.

\bibitem{a16}
X.~Zhu, W.~Su, L.~Lu, B.~Li, X.~Wang, and J.~Dai, ``Deformable detr: Deformable transformers for end-to-end object detection,'' 2021.

\bibitem{a17}
J.~Hu, L.~Cao, Y.~Lu, S.~Zhang, Y.~Wang, K.~Li, F.~Huang, L.~Shao, and R.~Ji, ``Istr: End-to-end instance segmentation with transformers,'' 2021.

\bibitem{a18}
H.~Wang, Y.~Zhu, H.~Adam, A.~Yuille, and L.-C. Chen, ``Max-deeplab: End-to-end panoptic segmentation with mask transformers,'' 2021.

\bibitem{a6}
Z.~Liu, Y.~Lin, Y.~Cao, H.~Hu, Y.~Wei, Z.~Zhang, S.~Lin, and B.~Guo, ``Swin transformer: Hierarchical vision transformer using shifted windows,'' pp.~9992--10002, 2021.

\bibitem{a7}
G.~Wu, W.-S. Zheng, Y.~Lu, and Q.~Tian, ``Pslt: A light-weight vision transformer with ladder self-attention and progressive shift,'' {\em IEEE Transactions on Pattern Analysis and Machine Intelligence}, vol.~45, no.~9, pp.~11120--11135, 2023.

\bibitem{a8}
X.~Liu, H.~Peng, N.~Zheng, Y.~Yang, H.~Hu, and Y.~Yuan, ``Efficientvit: Memory efficient vision transformer with cascaded group attention,'' in {\em 2023 IEEE/CVF Conference on Computer Vision and Pattern Recognition (CVPR)}, IEEE, June 2023.

\bibitem{a9}
S.~Mehta and M.~Rastegari, ``Mobilevit: Light-weight, general-purpose, and mobile-friendly vision transformer,'' 2021.

\bibitem{a10}
Z.~Dai, H.~Liu, Q.~V. Le, and M.~Tan, ``Coatnet: Marrying convolution and attention for all data sizes,'' 2021.

\bibitem{a11}
C.~Yang, S.~Qiao, Q.~Yu, X.~Yuan, Y.~Zhu, A.~Yuille, H.~Adam, and L.-C. Chen, ``Moat: Alternating mobile convolution and attention brings strong vision models,'' 2023.

\bibitem{a12}
S.~Wang, J.~Gao, Z.~Li, X.~Zhang, and W.~Hu, ``A closer look at self-supervised lightweight vision transformers,'' 2023.

\bibitem{a19}
W.~Wang, R.~Jiang, N.~Cui, Q.~Li, F.~Yuan, and Z.~Xiao, ``Semi-supervised vision transformer with adaptive token sampling for breast cancer classification,'' {\em Frontiers in Pharmacology}, vol.~13, p.~929755, 2022.

\bibitem{a20}
S.~Tummala, J.~Kim, and S.~Kadry, ``Breast-net: Multi-class classification of breast cancer from histopathological images using ensemble of swin transformers,'' {\em Mathematics}, vol.~10, no.~21, 2022.

\bibitem{a21}
G.~Baroni, L.~Rasotto, K.~Roitero, A.~Tulisso, C.~di~loreto, and V.~Della~Mea, ``Optimizing vision transformers for histopathology: Pretraining and normalization in breast cancer classification,'' {\em Journal of Imaging}, vol.~10, p.~108, 04 2024.

\bibitem{a24}
K.~Simonyan and A.~Zisserman, ``Very deep convolutional networks for large-scale image recognition,'' 2015.

\bibitem{a25}
K.~He, X.~Zhang, S.~Ren, and J.~Sun, ``Deep residual learning for image recognition,'' 2015.

\end{thebibliography}

\end{document}